\documentclass{article}

\PassOptionsToPackage{numbers, sort}{natbib}

    \usepackage[preprint]{neurips_2024}

\usepackage{amsthm}
\newtheorem{theorem}{Theorem}

\usepackage{graphicx} 
\usepackage{epsfig}
\usepackage{stfloats}
\usepackage[utf8]{inputenc} %
\usepackage[T1]{fontenc}    %
\usepackage[table]{xcolor}
\definecolor{citecolor}{HTML}{0071bc}
\usepackage[pagebackref,breaklinks=true,letterpaper=true,colorlinks,citecolor=citecolor,bookmarks=false]{hyperref}
\usepackage{url}            %
\usepackage{booktabs}       %
\usepackage{amsfonts}       %
\usepackage{nicefrac}       %
\usepackage{microtype}      %
\usepackage{xcolor}         %
\usepackage{graphicx}
\usepackage{amsmath}
\usepackage{algorithm}
\usepackage{comment}
\usepackage{amssymb}
\usepackage[noend]{algpseudocode}
\usepackage{bbm}
\usepackage{dsfont}
\usepackage{easylist}
\usepackage{xspace}
\usepackage{wrapfig}
\usepackage{multirow} 
\usepackage{amssymb}
\usepackage{subcaption}
\usepackage{bbding}
\usepackage{algorithm}      
\usepackage{algpseudocode}  

\usepackage{graphicx, amsmath, amssymb, caption, subcaption, multirow, overpic, textpos}
\usepackage{colortbl}
\usepackage[british, english, american]{babel}

\renewcommand{\paragraph}[1]{\vspace{1.25mm}\noindent\textbf{#1}}

\newcolumntype{x}[1]{>{\centering\arraybackslash}p{#1pt}}
\newcolumntype{y}[1]{>{\raggedright\arraybackslash}p{#1pt}}
\newcolumntype{z}[1]{>{\raggedleft\arraybackslash}p{#1pt}}
\newcommand{\app}{\raise.17ex\hbox{$\scriptstyle\sim$}}

\definecolor{deemph}{gray}{0.6}

\definecolor{baselinecolor}{gray}{.9}

\definecolor{lightcolor}{RGB}{137, 207, 240}

\newcommand{\remove}[1]{}

\makeatletter
\DeclareRobustCommand\onedot{\futurelet\@let@token\@onedot}
\def\@onedot{\ifx\@let@token.\else.\null\fi\xspace}

\title{Rectified Flow For Structure Based Drug Design}

\author{%
  Daiheng Zhang $^{1}$ \thanks{Daiheng and Chengyue contributed equally to this work.}\and Chengyue Gong $^{2}$ \and Qiang Liu$^{2}$ \\
\AND
   \textnormal{$^1$Department of Electrical and Computer Engineering,  Rutgers University}\\
 \textnormal{$^2$Department of Computer Science, The University of Texas at Austin}
  \\ 
}

\begin{document}

\maketitle

\begin{abstract} 
Deep generative models have achieved tremendous success in structure-based drug design in recent years, 
especially for generating 3D ligand molecules that bind to specific protein pocket. 
Notably, diffusion models have transformed ligand generation by providing exceptional quality and creativity. However, traditional diffusion models are restricted by their conventional learning objectives, which limit their broader applicability. 
In this work, we propose a new framework \textbf{FlowSBDD}, which is based on rectified flow model, allows us to flexibly incorporate additional loss to optimize specific target and introduce additional condition either as an extra input condition or replacing the initial Gaussian distribution.
Extensive experiments on CrossDocked2020 show that our approach could achieve state-of-the-art performance on generating high-affinity molecules while maintaining proper molecular properties without specifically designing binding site, 
with up to -8.50 Avg. Vina Dock score and 75.0\% Diversity.

\end{abstract}

\section{Introduction} 

In recent years, deep generative models have made substantial advancements in structure-based drug design~\citep{polykovskiy2020molecular}, especially in identifying lead molecules for known protein binding pocket. Various methodologies have been employed to generate ligands corresponding to these protein pockets. These include 1D representations like SMILES strings~\citep[\emph{e.g.}, ][]{bjerrum2017molecular,kusner2017grammar}, 2D graph representations~\citep{jin2018junction}, and 3D structures~\citep{gebauer2018generating,simm2020symmetry}. Each form offers unique insights, with 3D structures being particularly advantageous due to their ability to capture the intricate spatial relationships for understanding molecular interactions, which makes structure-based drug design ~\citep{anderson2003process} stands at the forefront of computational approaches to drug discovery. 

Structure-based drug design (SBDD) can be formulated as a conditional generation in a protein pocket micro-environment. Several deep generative models achieved promising results in SBDD field. Autoregressive and non-autoregressive are two commonly used models. In the previously proposed methods ~\citep[\emph{e.g.},][]{luo20213d,peng2022pocket2mol,liu2022generating,guan20233d, guan2023decompdiff}, autoregressive methods achieved better performance on the molecular properties~\citep{peng2022pocket2mol}, while diffusion approaches achieved better performance on the affinity metrics~\citep{guan20233d, guan2023decompdiff}.

We propose to approach SBDD with rectified flow~\cite{liu2022rectified, liu2022flow}, a powerful flow-based generative model framework. \textbf{FlowSBDD} allows us to construct a flexible generative model that can produce atom types and atom positions based on the given protein pocket information. Additionally, we propose a novel bond distance loss, which could enhance the learning process and model performance effectively. 
Moreover, we replace the standard gaussian initialization with other random distribution and well-generated molecules, and demonstrate that our framework can learn a model which refine the existing distribution. Empirically, our model can generate ligand molecules with the new SOTA \textbf{-8.50 Avg. Vina Dock Score} and \textbf{75\% Avg. Diversity} on the CrossDocked2020 \citep{francoeur2020three} benchmark.

\section{Methods} 


\subsection{Problem Definition}
\label{sec:prb_def}

We aim to generate ligand molecules that are capable of binding to specific protein binding sites, by modeling \( p(M|P) \). 
We represent the protein pocket as a collection of \( N_P \) atoms, \( P = \{ (x^{i}_P, v^{i}_P) \}_{i=1}^{N_P} \). Similarly, the ligand molecule can be represented as a collection of \( N_M \) atoms, \( M = \{ (x^{i}_M, v^{i}_M) \}_{i=1}^{N_M} \), where \( x^i_M \in \mathbb{R}^3 \) represents atom position and \( v^{i}_M \in [k] \)  the $k$ possible atom types. 
The number of atoms \( N_M \) can be sampled from an empirical distribution \citep{hoogeboom2022equivariant,guan20233d}. For brevity, the ligand molecule is denoted as \(\ M = \{ \mathbf{X}, \mathbf{V} \} \) where  \( \mathbf{X} \in \mathbb{R}^{N_M \times 3} \) and \( \mathbf{V} \in [k]^{N_M \times K} \).

\subsection{Rectified Flow}
\label{sec:method}

Rectified Flow \citep{liu2022rectified,liu2022flow}
provides an approach for learning a transport mapping between two distributions. 
We use $M_0$ to denote the initializing state, use $M_1$ to denote the final state, $M_t$  is the interpolated state at time \(t\). Rectified Flow learns to transfer $M_0$ to $M_1$
via an ordinary differential equation (ODE), or flow model
    $\mathrm{d}M_t / \mathrm{d}t = v(M_t, t), \text{ with } t \in [0, 1]$,
where \(\mathrm{d}t\) denotes the derivative with respect to time. And the time parameter $t$, which ranges from 0 to 1, denotes the progression from the initial to the final state.
Here $M_t$ is modeled by a velocity field $v (M_t, t)$. 
So we can construct a neural network to learn the velocity field $v (M_t, t)$ . The optimal trajectory at any given time \( t \) is directed from $M_0$ to $M_1$. Therefore, we guide our velocity field to closely adhere to the ideal linear path pointing from $M_0$ to $M_1$ as much as possible, by optimizing the following objective:
\begin{equation}
\label{traing_equa}
\min_\theta \mathbb{E}_{t \sim [0, 1]}\mathbb{E}\left\| v_\theta(M_t, t) - (M_1 - M_0) \right\|^2,
\end{equation}
where \( M_t = tM_1 + (1 - t)M_0 \).
This approach is employed for training both atom positions and atom types. It is noteworthy that atom types are constructed as one-hot vectors containing information such as element types and membership in an aromatic ring. The final loss is a weighted sum of atom coordinate loss and atom type loss. 

In practice, we can estimate the neural velocity field by Euler solver in \( N \) steps,
{\small
\begin{equation}
\label{sample_equa}
M_{t+\frac{1}{N}} = M_t + \frac{1}{N}v(M_t, t), \quad \forall t \in \left\{0, \ldots, N-1\right\}/N, 
\end{equation}
}where we complete the simulation with \( N \) steps. Furthermore, considering the extensive number of steps required by diffusion models to generate high-quality results, it is intuitive that the Euler solver's accuracy improves with a larger number of steps \( N \). Considering large \( N \) causing high computational cost, in our subsequent experiments, we set the \( N \) to a constant value that balances both generation quality and efficiency.

\paragraph{Rectified Flow with Bond Loss}
\label{para:additional-loss}
Considering the flexibility of the flow model, we can further utilize different metrics to enhance the model learning ability.
Here, we propose to apply a function which measures the difference in bond distances or angles.
The optimization approach is theoretically supported by an theorem \citep{liu2022rectified}, given by:

\begin{theorem}
{\bf(Informal)} For any user-specified given cost function \( c \), the rectified flow \( dZ_t = v(Z_t) dt \) can find an optimal coupling between the source and target distributions.
\end{theorem}
We refer the readers to ~\cite{liu2022rectified} for more formal conclusions, 
In this work, we introduce a bond-specific loss function as part of our model's optimization process, targeting the preservation of bond lengths, which could be formulated as:
\begin{equation}
\label{equa:bond}
\min_{\theta} \mathbb{E} \left[ \|\phi(M_{1}) - \phi(M_{0} + v(M_t, t))\|^2 \right].
\end{equation}

In \eqref{equa:bond}, 
\( M_1 \) represents the true ligand features, and \( M_0 \) is the initial state of the ligand features. The function \( \phi \) represents the metric and the velocity field \( v \) is parameterized by \( \theta \). 
For each pair of atoms that form a bond, \( \phi \) measures the difference between different example in the Euclidean space. More specifically, we calculate the loss by
\begin{equation}
\min_\theta \mathbb{E}_{M_0, M_1 \sim \mathcal{D}}   \sum_i \|  b_i(M_0) - b_i(M_1) \|, 
\label{eq:bond}
\end{equation}
where \( b_i \) represent function to extract the atom positions which forming the \( i \)-th bond. The summation over \( i \) ensures that the bond loss is computed over all bonds in the ligand. Finally, we sum \eqref{eq:bond} and \eqref{traing_equa} to formulate our training loss.


\paragraph{Flexible prior distribution}
\label{para:additional_conditions}
With the rectified flow, one advantage is that we can flexibly add additional prior to the model input. We can instead utilize a flexible initial distribution. Firstly, instead of relying on a standard Gaussian distribution, we can employ a random non-standard distribution.
We refer the readers to \cite{liu2022rectified} for more detailed discussions about this ability. 
Secondly, we can also transform from a data distribution which is generated by Molecular dynamics (MD) simulations \citep{hollingsworth2018molecular} or other deep learning models. In practice, we used the generated result from SOTA model \textbf{TargetDiff}.

\section{Experiments}


\subsection{Experimental Setup}
\label{sec:Experimental_Setup}
\paragraph{Datasets}
Our experiments were conducted using the CrossDocked2020 dataset  \citep{francoeur2020three}. Consistent with prior studies { \citep[e.g.,][]{luo20213d,peng2022pocket2mol,guan20233d,guan2023decompdiff} }, we adhered to the same dataset filtering and partitioning methodologies. We further refined the 22.5 million docked protein binding complexes, characterized by an RMSD $<  1 \AA $, and sequence identity less than 30\%. This resulted in a dataset comprising 100,000 protein-binding complexes for training, alongside a set of 100 novel complexes designated for testing.

\paragraph{Model architecture}
Inspired from recent progress in equivariant neural networks \citep{satorras2021n}, 
we model the interaction between the ligand molecule atoms and the protein atoms with a SE(3)-Equivariant GNN, the atom hidden embedding and coordinates are updated alternately in each layer, which follows \cite{guan20233d}.

\paragraph{Baseline} For benchmarking, we group various baselines according to their generative approaches: GNN-based Auto-regressive models such as \textbf{AR} \citep{luo20213d}, \textbf{Pocket2Mol} \citep{peng2022pocket2mol}, and \textbf{GraphBP} \citep{liu2022generating}; and other generative models like \textbf{liGAN} \citep{ragoza2022generating}, which is a 3D CNN-based method following a conditional VAE scheme. Except auto-regressive models, starting from vae/gan, researchers dive in diffusion based method, with \textbf{TargetDiff} \citep{guan20233d} and \textbf{DecompDiff} \citep{guan2023decompdiff} achieving notable results.

\paragraph{Evaluation} 
We evaluate generated ligands from three aspects: \textbf{target binding affinity and molecular properties}, \textbf{molecular structures} and \textbf{sampling efficiency}. For target binding affinity and molecular properties, we present our results under the best setting as \textbf{ours}. Following previous work~\citep{luo2021diffusion,ragoza2022generating}, we utilize AutoDock Vina~\citep{eberhardt2021autodock} for binding affinity estimation. We collect all generated molecules across 100 test proteins, with metrics reported for \textbf{target binding affinity} (Vina Score, Vina Min, Vina Dock, and High Affinity) and \textbf{molecular properties} (QED, SA, Diversity). 

\begin{table}[!ht]
\centering
\scalebox{0.75}{
\begin{tabular}{c|cc|cc|cc|cc|cc|cc|cc}
\toprule
 Model  & \multicolumn{2}{c|}{Vina Score ($\downarrow$)} & \multicolumn{2}{c|}{Vina Min ($\downarrow$)} & \multicolumn{2}{c|}{Vina Dock ($\downarrow$)} & \multicolumn{2}{c|}{High Affinity ($\uparrow$)} & \multicolumn{2}{c|}{QED ($\uparrow$)}   & \multicolumn{2}{c|}{SA ($\uparrow$)} & \multicolumn{2}{c}{Diversity ($\uparrow$)} \\
 & Avg. & Med. & Avg. & Med. & Avg. & Med. & Avg. & Med. & Avg. & Med. & Avg. & Med. & Avg. & Med.  \\
\midrule
\multicolumn{15}{c}{\textbf{Comparison with Autoregressive Methods}} \\
\midrule
liGAN $^*$       & - & - & - & - & -6.33 & -6.20 & 21.1\% & 11.1\% & 0.39 & 0.39 & 0.59 & 0.57 & 0.65 & 0.70 \\
GraphBP $^*$     & - & - & - & - & -4.80 & -4.70 & 14.2\% & 6.7\% & 0.43 & 0.45 & 0.49 & 0.48 & \underline{0.79} & \underline{0.78} \\
AR          & \textbf{-5.75} &\textbf{ -5.64} & -6.18 & \underline{-5.88} & -6.75 & -6.62 & 37.9\% & 31.0\% & \underline{0.51} & \underline{0.50} & \underline{0.63} & \underline{0.63} & 0.70 & 0.74 \\
Pocket2Mol  & \underline{-5.14} & -4.70 & \underline{-6.42} & -5.82 & \underline{-7.15} & \underline{-6.79} & \underline{48.4\%} & \underline{51.0\%} & \textbf{0.56} & \textbf{0.57} & \textbf{0.74} & \textbf{0.75} & 0.69 & 0.72 \\
FlowSBDD & -3.62& \underline{-5.03} &\textbf{-6.72} &\textbf{-6.60} & \textbf{-8.50} & \textbf{-8.36} & \textbf{63.4\%} & \textbf{70.9\%} & 0.47 & 0.48 &0.51 & 0.51& 0.75 & 0.75\\

\midrule
\multicolumn{15}{c}{\textbf{Comparison with Probabilistic Flow Methods}} \\
\midrule
TargetDiff  & \underline{-5.47} & \textbf{-6.30} & -6.64 & \underline{-6.83} & -7.80 & -7.91 & 58.1\% & 59.1\% & \textbf{0.48} & \textbf{0.48} & 0.58 & 0.58 & 0.72 & 0.71 \\
DecompDiff & \textbf{-5.67}& \underline{-6.04} & \textbf{-7.04} & \textbf{-7.09} & \underline{- 8.39} & \textbf{-8.43} & \textbf{64.4\%} & \textbf{71.0\%} & 0.45 &  \underline{0.43}& \textbf{0.61}& \textbf{0.60}& 0.68 & 0.68 \\
FlowSBDD & -3.62& -5.03 &\underline{-6.72} &-6.60 & \textbf{-8.50} & \underline{-8.36} & \underline{63.4\%} & \underline{70.9\%} &  \underline{0.47} & \textbf{0.48} &0.51 & 0.51 & \underline{0.75} & \underline{0.75}\\
\midrule
Reference   & -6.36 & -6.46 & -6.71 & -6.49 & -7.45 & -7.26 & -  & - & 0.48 & 0.47 & 0.73 & 0.74 & -    & -    \\
\bottomrule
\end{tabular}}
\vspace{10pt}
\caption{Summary of different properties of reference molecules and molecules generated by our model and other baselines. (↑)/(↓) denotes a larger/smaller number is better. Top 2 results are highlighted with bold text and underlined text, respectively. }
\label{tab:main_result}
\end{table}

\begin{figure*}[!ht]
    \centering
    \includegraphics[width=0.9\linewidth]{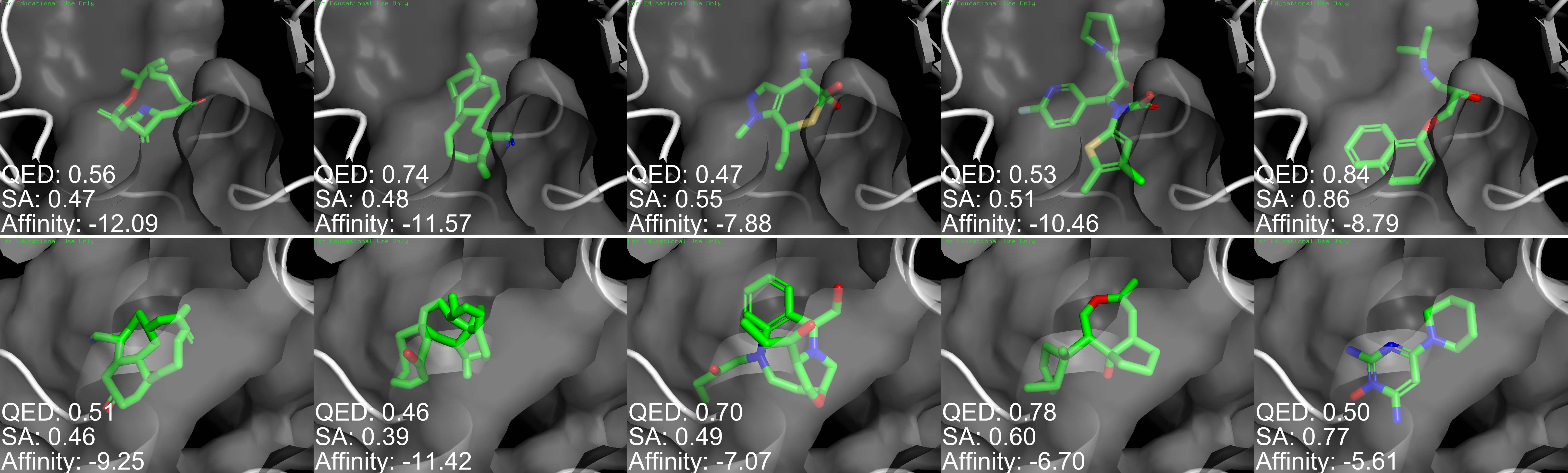}
    \caption{Visualization of FlowSBDD in best setting(left two columns), TargetDiff(middle two columns), and reference binding molecules (right column) on protein with PDB code 2V3R (top row) and 3B6H (bottom row). We apply mesh representation and notice that the molecule is docked well.}
    \label{fig:vis different methods}
\end{figure*}
\subsection{Rectified Flow is an alternative choice for SBDD}
\label{sec:result}
\begin{figure*}[!ht]
    \centering
    \includegraphics[width=1.0\linewidth]{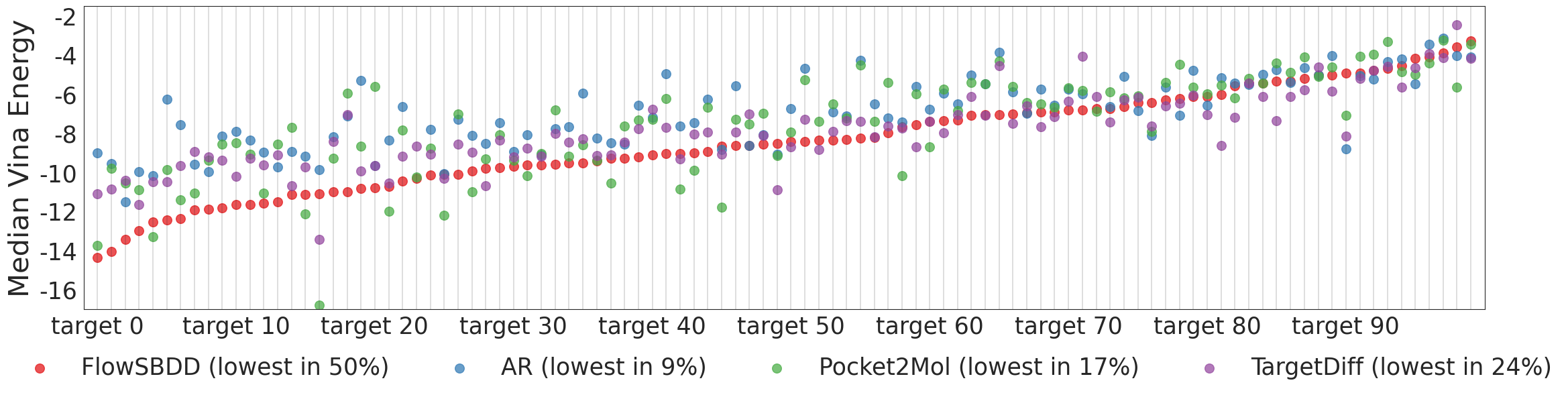}
    \vspace{-15pt}
    \caption{Median Vina energy for different generated molecules (AR vs. Pocket2mol vs. Targetdiff vs. FlowSBDD) across 100 testing binding targets.}
    \label{fig: Median_Vina}
\end{figure*}

\paragraph{Target binding affinity} Firstly, we evaluate the effectiveness of FlowSBDD in terms of binding affinity. As shown in Table \ref{tab:main_result}, we can see our model significantly outperforms autoregressive baselines in binding-related metrics. Compared with the state-of-the-art diffusion-based method \textbf{DecompDiff}, our model surpasses by 1.3\% in Avg. Vina Dock, and get second highest in Med. Vina Dock, which we consider as the most important factor in binding affinity. We also achieved the second highest Vina Min and High Affinity. This indicates that our model has the potential to generate molecules with better affinities with the pockets. We also show the median Vina energy {(computed by AutoDock Vina \citep{eberhardt2021autodock})} of all generated molecules for each binding pocket in Figure \ref{fig: Median_Vina}. Based on the Vina energy, generated molecules from our model show the best binding affinity in \textbf{50\%} of the targets, while the ones from \textbf{AR}, \textbf{Pocket2Mol} and \textbf{Targetdiff} are best for 9\%, 17\% and 24\% of all targets. 

\paragraph{Molecule property} Next, we evaluate molecular properties. Due to their step-by-step nature, autoregressive models are capable of generating better molecule properties. Here, we primarily compare FlowSBDD with other probability flow models. Our model achieved the best results in QED Med., Diversity Avg., and Med. The diversity exceeded the \textbf{Targetdiff} by 4.2\%. Compared to autoregressive models, it is noteworthy that our model's Diversity is only lower than \textbf{GraphBP} (with both molecule property and binding affinity being the lowest). This means that our model can generate drug-like molecules with higher diversity, which is a desirable scenario when generating drug candidates. The table \ref{tab:main_result} presents the result under the best setting.

\paragraph{Sampling efficiency} Here we show the time to generate 100 molecules for each pocket: AR took 7785 seconds, Pocket2Mol took 2544 seconds, and TargetDiff took 3428 seconds. GraphBP was notably faster at just 105 seconds, while DECOMPDiff took 6189 seconds. In contrast, FlowSBDD achieved a remarkable time of 144 seconds sampling 100 steps with vanilla Euler method. 

\subsection{Adaptive Enhancements Improving Performance} 

\paragraph{Enhancing Performance with Bond Euclidean Distance} As described in Section \ref{para:additional-loss}, we introduced a loss function based on capturing the bond Euclidean distance. Observations from the sampling results in Table \ref{tbl:add_bond}, demonstrate that the additional loss plays a significant role in enabling the model to learn better features, especially increasing the QED and Vina Dock. The Median QED improved by \( 17.6\% \), and the Avg. Vina Dock score showed an improvement by \( 29.4\% \).

\begin{table}[!ht]
\centering
\scalebox{0.8}{
\setlength{\tabcolsep}{4pt} 
\begin{tabular}{c|cc|cc|cc|cc|cc}
\toprule
Model  & \multicolumn{2}{c|}{Vina Score (↓)} & \multicolumn{2}{c|}{Vina Min (↓)} & \multicolumn{2}{c|}{Vina Dock (↓)} & \multicolumn{2}{c|}{QED (↑)}   & \multicolumn{2}{c}{SA (↑)}  \\
 & Avg. & Med. & Avg. & Med. & Avg. & Med. & Avg. & Med. & Avg. & Med.  \\
\midrule
\textit{Best w/o B} & -3.42 &-4.19& -5.12& -5.17 &-6.57& -6.89& 0.40 & 0.42 &0.60 & 0.59 \\
\textit{Best w/ B} & \textbf{-3.62}& \textbf{-5.03} & \textbf{-6.72}& \textbf{-6.60} & \textbf{-8.50} & \textbf{-8.36} & \textbf{0.47} & \textbf{0.48} & 0.51 & 0.51 \\

\bottomrule
\end{tabular}
}
\caption{Comparison of adding additional loss. \textit{Best w/o B} represents a standard Gaussian distribution scaled by factors of 0.001, with only atom position and atom type MSE loss. \textit{Best w/ B} represents a standard Gaussian distribution scaled by factors of 0.01 adding additional bond pairwise loss, which is also our best setting.
}
\label{tbl:add_bond}
\end{table}
\paragraph{Different noise level} We empirically study the effect of prior distribution which mentioned in \ref{para:additional_conditions}. Empirically, we configure the noise from two aspects: the type of noise, the scale of noise. 
As shown in table \ref{tbl:diff init}, we can observe that \(\textit{$\mathcal{N}(0, 0.001)$}\) achieved the best QED and Vina Dock results. More theoretical work is needed to explain this result. 

\paragraph{Starting from a `Better' Prior Distribution} We aim to explore a data-to-data generation process, to directly start from the result from a deep learning model. Due to the extensive time required to generate \textbf{Targetdiff} and the large size of the training set, we opted for a 100-step method when sampling to obtain the source distribution, instead of the original 1000-step approach. As shown in the table below, the results we generated are far superior to the initial 100-step \textbf{Targetdiff}. We believe that by using a better prior as the starting distribution, our model will achieve even better results. 

\begin{table}[!ht]
\centering
\scalebox{0.8}{
\setlength{\tabcolsep}{4pt} 
\begin{tabular}{c|cc|cc|cc|cc|cc}
\toprule
Model  & \multicolumn{2}{c|}{Vina Score (↓)} & \multicolumn{2}{c|}{Vina Min (↓)} & \multicolumn{2}{c|}{Vina Dock (↓)} & \multicolumn{2}{c|}{QED (↑)}   & \multicolumn{2}{c}{SA (↑)}  \\
 & Avg. & Med. & Avg. & Med. & Avg. & Med. & Avg. & Med. & Avg. & Med.  \\
\midrule
\textit{$\mathcal{U}(0, 1)$} & -3.53 & -4.20 & -4.73 & -4.79 & - & - &  0.33 & 0.31 & 0.59 & 0.58 \\
\textit{$\mathcal{N}(0, 1)$}  & -2.19 & -2.72 & -3.64 & -3.79 & - & - & 0.35 & 0.34 & \textbf{0.64} & \textbf{0.64} \\
\textit{$\mathcal{N}(0, 0.1)$} &  -2.17&-3.55 & -4.29&-4.41 & -5.28&-5.39& 0.36 &0.36& 0.61 & 0.60 \\
\textit{$\mathcal{N}(0, 0.01)$} & -3.28&-4.59 & -5.91&-5.89 & -7.48&-7.56 &0.40 &0.41 & 0.51 & 0.51 \\
\textit{$\mathcal{N}(0, 0.001)$} & \textbf{-3.62}& \textbf{-5.03} &\textbf{-6.72} &\textbf{-6.60} & \textbf{-8.50} & \textbf{-8.36} & \textbf{0.47} & \textbf{0.48} &0.51 & 0.51 \\
\textit{$\mathcal{N}(1, 1)$} & -2.75 &-4.28& -5.78& -5.59 & -7.24 & -7.20 &0.34 &0.33 & 0.48&  0.48 \\
\midrule
\textit{100Tar}  & 22.89& 10.64 & 3.44 & -2.56 & -4.04 & -4.48& 0.28 &0.26 & 0.36 & 0.35 \\
\textit{100tar-Flow} &\textbf{-2.57} & \textbf{-3.58} & \textbf{-5.10} & \textbf{-5.06} & \textbf{-6.69} & \textbf{-6.63} &\textbf{0.31}& \textbf{0.29} & \textbf{0.45} & \textbf{0.43} \\
\bottomrule
\end{tabular}
}
\caption{Summary of results of different prior distributions. We compare the effect of various noise distributions on the model's performance. Here, $\mathcal{U}(0, 1)$ represents a uniform distribution; $\mathcal{N}(0, 1)$, $\mathcal{N}(0, 0.1)$, $\mathcal{N}(0, 0.01)$, and $\mathcal{N}(0, 0.001)$ represent Gaussian distributions with mean 0 and standard deviations of 1, 0.1, 0.01, and 0.001 respectively; $\mathcal{N}(1, 1)$ represents a Gaussian distribution with mean 1 and standard deviation 1. \textit{100Tar} represents the results generated by \textbf{TargetDiff} using 100 euler steps, and also serves as our starting distribution. \textit{100tar-Flow} represents the results generated starting from the 100 euler steps \textbf{TargetDiff} using our method.}
\label{tbl:diff init}
\end{table}



\section{Conclusions} 
In this work, we propose to use a novel Rectified flow based generation framework \textbf{FlowSBDD} for the protein-specific molecule generation. Our framework can simply combine with adding additional bond loss function or changing the prior noise distribution, which achieves comparable performance to SOTA diffusion models while offering faster sampling speed and more flexible scaling options.

\paragraph{Limitations and Broader Impact} In this work, we propose a novel rectified flow-based framework \textbf{FlowSBDD} applied in Structure Based Drug Design. This framework demonstrates its good performance and flexibility, showcasing its potential in drug design and related area. In the future, we plan to continue exploring the impact of different priors and attempting to provide a theoretical explanation for this. We also found that our model \textbf{FlowSBDD} underperforms others on Vina Score metric, and we will investigate the unreasonable aspects of the generated structures in subsequent experiments.


{
\small
\bibliographystyle{splncs04}
\bibliography{main}
}





\newpage

\end{document}